\documentclass[sigconf]{acmart}
\usepackage{subcaption}
\usepackage{graphicx}

\begin{document}

\title{Detecting discriminatory risk through data annotation based on Bayesian inferences}

\author{Elena Beretta}
\email{elena.beretta@polito.it}
\orcid{0000-0002-6090-2765}
\affiliation{%
  \institution{Nexa Center for Internet \& Society \\ 
              Department of Control and Computer Engineering \\
              Politecnico di Torino}
  \streetaddress{Corso Duca degli Abruzzi, 24}
  \city{Turin}
  \country{Italy}
  \postcode{10129}\\
  \institution{Fondazione Bruno Kessler}
  \streetaddress{Via Sommarive, 18}
  \city{Trento}
  \country{Italy}
  \postcode{38123}
}

\author{Antonio Vetr\`{o}}
\email{antonio.vetro@polito.it}
\orcid{0000-0003-2027-3308}
\affiliation{%
  \institution{Nexa Center for Internet \& Society \\ 
              Department of Control and Computer Engineering \\
              Politecnico di Torino}
  \streetaddress{Corso Duca degli Abruzzi, 24}
  \city{Turin}
  \country{Italy}
  \postcode{10129}
}

\author{Bruno Lepri}
\email{lepri@fbk.eu}
\orcid{0000-0003-1275-2333}
\affiliation{%
  \institution{Fondazione Bruno Kessler}
  \streetaddress{Via Sommarive, 18}
  \city{Trento}
  \country{Italy}
  \postcode{38123}
}

\author{Juan Carlos De Martin}
\email{demartin@polito.it}
\orcid{0000-0002-7867-1926}
\affiliation{%
  \institution{Nexa Center for Internet \& Society \\ 
              Department of Control and Computer Engineering \\
              Politecnico di Torino}
  \streetaddress{Corso Duca degli Abruzzi, 24}
  \city{Turin}
  \country{Italy}
  \postcode{10129}
}

\renewcommand{\shortauthors}{Beretta et al.}

\begin{abstract}
Thanks to the increasing growth of computational power and data availability, the research in machine learning has advanced with tremendous rapidity. Nowadays, the majority of automatic decision making systems are based on data. However, it is well known that machine learning systems can present problematic results if they are built on partial or incomplete data. In fact, in recent years several studies have found a convergence of issues related to the ethics and transparency of these systems in the process of data collection and how they are recorded. Although the process of rigorous data collection and analysis is fundamental in the model design, this step is still largely overlooked by the machine learning community. For this reason, we propose a method of data annotation based on Bayesian statistical inference that aims to warn about the risk of discriminatory results of a given data set. In particular, our method aims to deepen knowledge and promote awareness about the sampling practices employed to create the training set, highlighting that the probability of success or failure conditioned to a minority membership is given by the structure of the data available. We empirically test our system on three datasets commonly accessed by the machine learning community and we investigate the risk of racial discrimination.
\end{abstract}



\maketitle

\section{Introduction}
\label{sec:intro}
In the last decades machine learning systems are widely spreading in different academic domains, as well as many public and private sectors are increasing the exploitation of these systems. Their widespread and pervasiveness is mainly driven by the exponential growth of computational power and the extensive availability of large amounts of data \cite{Kleinberg:18}. Supervised machine learning models are also particularly widespread and now deeply rooted in different sectors due to their usage versatility. The predictive ability of supervised machine learning systems is deployed in disparate areas of application: credit reliability \cite{Kleinberg:18}, justice system \cite{Angwin:2016, Berk:18}, job recommendations \cite{Siting:2012}, university selection process \cite{Kanoje:2016}, cultural contents \cite{Schedl:2018},\cite{Indira:2019} and purchases recommendations \cite{Oyebode:2020}.
The key ingredient that supervised machine learning models have in common is the availability of a set of labeled data used to train the model in elaborating a response related to past events \cite{Gebru:2018}. Since the known properties of the available set of data is used to create a classifier that makes predictions about new entities of the same type, the structure, properties and quality of the data are aspects that largely and directly influence the quality of the model and the results it produces \cite{ONeal:16}, \cite{Albarghouthi:2019}. Although data-driven decision models have been shown to produce both economic and social benefits, many researchers have highlighted several problems and damages related to their use in different areas, especially if they are built on partial or incomplete data \cite{Hardt:16}, \cite{Dwork:2012}.
As a matter of fact, in recent years several studies have found a convergence of issues related to the ethics and transparency of these systems in the process of data collection and in the way they are recorded \cite{Marda:2020}.
While the process of rigorous data collection and analysis is fundamental to the design of the model, this step is still largely overlooked by the machine learning community \cite{Beretta:19, Jo:2020}. As the practice of removing protected attributes from available data has been shown to potentially exacerbate further discrimination \cite{Williams:2018} - making bias even more difficult to detect - practices related to data collection, data transparency and data explainability become even more relevant and urgent. 
The aim of our work is to provide a data annotation system that serves as a diagnostic framework containing immediate information about the data appropriateness in order to more accurately assess the quality of the available data used in training models. We propose a data annotation method based on Bayesian statistical inference that aims to warn of the risk of discriminatory results of a given data set. In particular, our method aims to deepen the statistical knowledge related to the information contained in the available data, and to promote awareness of the sampling practices used to create the training set, highlighting that the probability of a discriminatory result is strongly influenced by the structure of the available data. We test our data annotation systems on three dataset widely spread in machine learning community: the COMPAS dataset \cite{Angwin:2016Compas}, the Drug Consumption dataset \cite{Fehrman:2015data}, \cite{Fehrman:2017paper} and the Adult dataset \cite{Kohavi:1996}.

\subsection{Problem Statement}
\label{sec:probStat}
The majority of machine learning systems are based on historical data processing \cite{Noble:2018}. This is particularly true in supervised machine learning models. Several studies have shown evidence that many equity and discrimination issues are due to input data properties \cite{Benjamin:2019}. Most of today data sets used to train models are chosen through non-probabilistic methods, generating problems of data imbalance and representativeness \cite{Eubanks:2018, Noble:2018}. This means that different fractions of the population do not show the same opportunity to be represented within the sample - aka, training sets -, leading some groups of individuals to have a lower probability of being represented. Common observed effects of a bad sampling are underestimation and overestimation of some groups \cite{Barocas:18}. Undetected distortions in data may also easily represent a spurious statistical noise. This happens when the data structure induces dependence between two variables that are not linked by a real cause-effect relationship.
\paragraph{\textbf{Data Sampling}} A key moment in the pipeline of a machine learning model is when the programmed algorithm is supplied with training data representing the entities on which the model itself trains its knowledge to make predictions. The quality of the data used in this phase is fundamental for the desired result, according to the principle of "garbage in - garbage out": even the most sophisticated models can present distorted results in the presence of low quality data \cite{Tommasi:17}. One of the main causes of data distortion is the way the data is selected and provided to the algorithm displaying problems related to inaccuracy, lack of update or inadequate representativeness. However, while knowledge of bias typologies has proliferated over the years, less attention is paid to issues concerning data collection, notation and sampling \cite{McDuff:2018}. In the spirit of fostering a broader awareness of data handling, we provide a reasoned list of issues that may arise during this phase:
\begin{itemize}
    \item [i] \textit{Data Selcetion:} the large proliferation of data sets availability on the same kind of problem to be analyzed, make hard the a priori choice of a given data set \cite{Gebru:2018};
    \item [ii] \textit{Inadequate sampling methods:} most models are trained with data sets that have been "found" and not subjected to probabilistic sampling methods, leading to limited or no data control \cite{Asudeh:2019Data};
    \item [iii] \textit{Cost and Time Limit:} collecting large amounts of data that present proportional representations of each property with respect to a sensitive attribute is time consuming and often costly and labor-intensive \cite{Caliskan:2017};
    \item [iv] \textit{Data set validation:} in the design of a machine learning model, more attention is paid to the mathematical basis of the classifier, restricting the data formation process to a black box \cite{Gil:2016, Geiger:2020};
    \item [v] \textit{Validation planning:} data validation, when applied, is often performed only after the model has been trained and used, making the feedback cycle inefficient and often ineffective \cite{Holland:2018};
    \item [vi] \textit{Lack of statistical rigorousness:} the suitability of the data set varies depending on the task for which the data are prepared. For instance, models based on linear regression imply assumptions of normality on the measurement error \cite{Gebru:2018, Wang:2019}. This specificity is often absent in the pipeline of machine learning models.
\end{itemize}
\paragraph{\textbf{Miss-dependency}} Two-dimensional or bivariate statistics is the study of the degree to which two distinct characters of the same statistical unit are connected. However, the connection only measures the degree of statistical dependency without inducing a cause-effect relationship or dependency between the variables. For instance, it can be shown that people with small feet make more spelling mistakes than people with large feet. However, this statistical dependency does not indicate that having small feet is the cause of spelling mistakes; the greater frequency of spelling mistakes may in fact be due to the younger age of people with small feet. In this case there could be a third variable, age, responsible for the cause-effect relationship. While in a human-centered model - where the human makes the decisions - this distinction is quite evident, in a machine learning model miss-dependency is not always deductible. This depends on two reasons: i) the machine does not recognize the meaning of the instance but looks at the properties of the variables; ii) the way in which the data are structured modifies the interpretation that the machine is having regarding the relation of statistical dependency. This means that, while in a human-centered model it is the human to verify that the relationships of statistical dependence detected in the available data are leading to a cause-effect relationship, in machine learning models the machine is not always able to recognize a spurious connection, erroneously assigning to two or more variables a cause relationship. In other words, the structure of the available data is responsible for the successful or failed relationships established with the protected attributes (ethnicity, gender, etc.) in the data. In addition, the rapid growth and spread of current machine learning systems is due in part to the ease of design of the models themselves, which thanks to modern software allows the construction of predictive models avoiding the understanding and adoption of rigorous statistical analysis. The simplicity of design has therefore created a gap between predictive and analytical-explicative power, favoring misinterpretation between causality and statistical dependence. The distinction between statistical dependence and causal dependence in data is therefore a primary issue in machine learning models, especially to determine the causes of failure, potential biases encoded in the data and the reliability of application. \\
Based on the problems highlighted, our contribution aims to answer the following research questions:
\begin{itemize}
    \item [\textbf{RQ1}] Is it possible to establish the probability of composition of the training data from the available data set?
    \item [\textbf{RQ2}] Do the available data known to the machine learning community present a discriminatory future risk based on their structure?
\end{itemize}

\section{Background}
\label{sec:background}
When machine learning model decisions are based on historical records, they tend to embed distortions that exist in reality and crystallize them. Prejudices and human bias therefore become part of the technology itself. This is particularly evident with regard to ethnic discrimination. Over the last years, the rise of machine learning models in various sectors is leading to a dramatic increase of discriminatory outcomes for ethnic minorities, across different fields of application. A striking and well known case is the COMPAS software, used in U.S. court to estimate the probability of defendants' recidivism, which has been shown to underestimate the risk of recidivism for white defendants and overestimate it for black defendants \cite{Angwin:2016Compas}. However, the COMPAS case is not an isolated phenomenon. In a 2017 experiment conducted on the Airbnb platform, applications from guests with typically African American names were found to be 16\% less likely to be accepted than identical guests with typically white names \cite{Edelman:2017}. Also in 2017, a geo-statistical analysis revealed that the design of the popular Pokémon GO game strengthens existing geographical prejudices, for example by benefiting urban areas and neighborhoods with smaller minority populations, economically disadvantaging ethnic minority areas \cite{Colley:2017}. Several studies have demonstrated the discriminatory potential of targeting advertising \cite{Speicher:2018}, \cite{Song:2020}, which is only recently receiving interventions to remove the prejudicial content of the model. For example, Facebook after years of scandals related to ads that exclude people based on race \cite{Angwin:2016} has finally removed the racial targeting option for ads \cite{Kukura:2020}. In a 2019 study, the commercial algorithm widely used in the U.S. health care system to guide health care decisions was found to discriminate against black patients \cite{Obermeyer:2019}. The algorithm falsely assigned a healthier condition to black patients despite the risk of complications being the same for white patients, making black people less likely to receive more financial resources for extra care.
Although facial recognition technologies are now used in several domains, they still present many discriminatory issues related to differences in margins of error - generally software has a 20\% higher margin of recognition error for black women \cite{Raji:2020} -. As an example, we report what happened recently with Google Vision AI, a computer vision service for image labeling \cite{Kayser:2020}. By providing the system with two images of people holding a body temperature thermometer, it labeled the image containing the white person as an "electronic device", while in the image containing the black person the device held was labeled as a "gun". In a later experiment it was shown that it was sufficient to apply a pink mask on the black person's hand in order the software labeled the image as "tool". 
Racial bias encoded in machine learning systems is likely to spread silently and like wild fire in everyday technologies. The increasing and ubiquitous spread of such models also intended to make allocative decisions about people's lives makes the problem of prejudice and rational discrimination more urgent than ever. For this reason and for the historical moment we are experiencing, our work intends to focus on rational discrimination in data.

\section{Motivating Example}
\label{sec:motEx}
\textit{Given a population composed of 60\% Caucasians, 35\% black people and 15\% Asian people, the probability of positive outcome for the respective ethnic groups is 70\% for Caucasians, 20\% for Blacks and 60\% for Asians. What is the probability of failure with respect to the protected attribute Ethnicity?} 

In this example the probabilities are given rather than the numerosity in order to simplify the following notation. To offer a better a better understanding of the Methodology this data will be used in Section \ref{sec:methodology}. The data gives the probability of success, but the similar reasoning is also valid for cases where the probability of failure is known. The intent is to verify whether the probabilities of success or failure of a subgroup are influenced by group membership - and vice versa - and more specifically how these probabilities affect the composition of the training set.

\section{Methodology}
\label{sec:methodology}
Our data annotation system is based on four modules:
\begin{itemize}
    \item [\textbf{I}] \textbf{Dependence}: assesses the degree of connection among the protected attribute - in our study, the ethnicity - and the target variable;
    \item [\textbf{II}] \textbf{Diverseness}: provides the training diversification probability in respect to each level of the protected attribute and the target variable;
    \item [\textbf{III}] \textbf{Inclusiveness}: provides the probability that two properties are simultaneously included in the training set;
    \item [\textbf{IV}] \textbf{Training Likelihood}: provides the occurrence likelihood of the protected attribute levels given the target variable levels - and vice versa - before the training set is sampled. 
\end{itemize}
\subsection{Quantifying Dependence}
\label{sec:dependence}
Excluding some specific domains where the dependence of some protected attributes with the response variable is not considered problematic, but rather it is fundamental for the understanding of a certain problem (for example the gender attribute in the medical field in the detection of particular diseases \cite{Dahiwade:2019}), in the broad field of machine learning systems the dependence between the protected attribute and the response variable has caused severe consequences \cite{ONeal:16, Noble:2018}. The dependence between the protected attribute and the response variable is therefore one of the major causes of discrimination and as such must be rigorously examined. The first step for a correct bias detection within the data is given by the dependency analysis between the different modalities of a protected attribute and the response variable.
In statistics, the measurement of the degree of dependence of two qualitative variables is called contingency; contingency measures the degree of connection of two categorical variables. To determine the degree of connection, the marginal frequencies and the combined frequencies of the bivariate table are used. Given two categorical variables \(x_i\) and \(y_i\), the dependency or independence is established through the theoretical independence table \(f'(x_i,y_j)\) once the table of the observed real data \(f'(x_i,y_j)\) is given. The contingency \(C(x_i;y_j)\) is therefore given by the difference between the observed and theoretical frequencies:
\begin{equation}\label{eq:contingency}
    C(x_i;y_j) = f(x_i,y_j) - f'(x_i,y_j) 
\end{equation}
If the table of the observed real data and the theoretical table of independence coincide - that is if for each cell the value is null - then the two variables are independent. Otherwise, it is necessary to measure the degree of connection between the variables.
The degree of connection between two categorical variables is commonly measured by the Pearson connection index, obtained as the sum of the relative quadratic contingencies. The index assumes a value of zero in case of independence in distribution and increases as the degree of connection between variables increases:
\begin{equation}\label{eq:pearson}
    \chi^2 = \sum_{i,j}\frac{C^2(x_i;y_j)}{n_{i,j}} = n \Big(\sum_{i,j}\frac{n_{i,j}^2}{n_{i0}n_{0j}}\Big)
\end{equation}
In order to support Pearson's connection index, the contingency coefficient is adopted with the purpose of reducing the \(\chi^2\) in the range [0;1]:
\begin{equation}\label{eq:contCoeff}
    C = \sqrt{\frac{\chi^2}{\chi^2+n}}
\end{equation}
However, the effect size of the degree of connection between two categorical variables is not always easy to interpret, where by effect size we mean a quantitative measure of the magnitude of a phenomenon. To offer a better understanding of the relationship of dependency between two variables, several simplified methods of interpretation have been proposed, especially to guide social scientists in the interpretation of statistical test results. In the spirit of simplifying the interpretation of the dependency between the response variable and the protected categories for a data set user, we introduce the concept of the Effect Size Index w (ES w):
\begin{equation}\label{eq:ESw}
    w = \sqrt{\sum_{i=1}\frac{(P_{1i}-P_{0i})^2}{P_{0i}}},
\end{equation}
where \(p_0i\) and \(p_1i\) are the value of the ith cells.
Notice that unlike the contingency coefficient, the ES w is not derived from frequencies but from proportions. The relationship between the Pearson connection index, the contingency coefficient and the ES Index is given by the following formula:
\begin{equation}\label{eq:relation}
    C = \sqrt{\frac{\chi^2}{\chi^2+n}} = \sqrt{\frac{w^2}{w^2+1}}
\end{equation}
Alternatively to the Formula \ref{eq:ESw} it is also possible to calculate the ES w from the contingency coefficient:
\begin{equation}
    w = \sqrt{\frac{C^2}{1-C^2}}
\end{equation}
The size of the ES w between two variables is then evaluated through the use of Table \ref{tab:esInter}, which relates the magnitude of the ES with a nominal label.
\begin{table} [ht!]
\caption{Conventional definitions of Effect Size Index w magnitude}
\label{tab:esInter}
\begin{tabular}{ll}
\toprule
\textbf{Magnitude} & \textbf{Value}\\
\hline
SMALL & w = 0.1\\
MEDIUM & w = 0.3\\
LARGE & w = 0.5\\
\bottomrule
\end{tabular}%
\end{table}
The advantage of using the conventional conversion table for the user of the data set is that the magnitude of the dependency is displayed quickly and immediately without the need for more complex statistical tests.

\subsection{Estimating Diverseness}
\label{sec:diverseness}
Intuitively, the probability of an event represents how likely the event will occur. According to the classical definition the probability is given by the following ratio:
\begin{equation}
    P = \frac{number\: of\: favorable\: cases}{number\: of\: possible\: cases}
\end{equation}
We now apply this elementary theory to the problem of data collection in machine learning. When the data set is partitioned into training and test sets, a split with a more or less standard ratio (70/30 or 80/20) is generally performed, i.e. a sampling is performed on the available data. Let's consider the case in which the training data set is generated by random sampling on the original data set without considering further techniques (stratification or re-sampling) - for example in the case of a non expert user -. The probability an event occurs turns into the probability that the training set shows some existing properties contained in the original data set:
\begin{equation}
    P = \frac{number\: of\: favorable\: properties}{number\: of\: possible\: properties}
\end{equation}
In our data annotation this ratio is introduced to allow the dataset user to answer questions like: \textit{"If I perform a random sampling on the original dataset, what is the probability that the training set is mainly composed of positive examples? What is the probability of belonging to a certain group with respect to the target variable?"}
\paragraph{\textbf{Prior Probabilities}} The a priori probability of a data property is the degree of belief of the property in the absence of other information, also known as the unconditional probability. The degree of belief is the probability of a property to be true in an uncertain environment. The probability is referred to the belief and not to the truth of the fact, as it is not possible for the user to know exactly the truth, that is if the original data are representative of the real world. Since the user does not have access to the complete information, several hypotheses on how the real data is structured have to be drawn, assigning to each of them a probability of being true. Formally:
\begin{equation}\label{eq:prior}
\begin{split}
    P &= (Y = y) \\
    P &= (A = a)
\end{split}
\end{equation}
We estimate the prior probabilities by using the data of the problem introduced in Section \ref{sec:motEx}, where the target variable \(Y\) assumes value 1 in case of negative outcome, otherwise 0. 
\begin{table} [ht!]
\caption{Example of prior probabilities}
\label{tab:prior}
\begin{tabular}{ll}
\toprule
\textbf{Formula} & \textbf{Probability}\\
\hline
\(P(Y = 0)\) & P = 0.48\\
\(P(Y = 1)\) & P = 0.52\\
\hline
\(P(A = white)\) & P = 0.6\\
\(P(A = black)\) & P = 0.35\\
\(P(A = Asian)\) & P = 0.15\\
\bottomrule
\end{tabular}%
\end{table}
In this specific case, the prior probabilities indicate that the training set has probability 0.48 to be composed by individuals who display a positive outcome and 0.52 to be composed by individuals who display a negative outcome; finally, the probabilities that it is formed by individuals of white, black and Asian ethnicity are respectively 0.6, 0.35 and 0.15 (Table \ref{tab:prior}).

\subsection{Estimating Inclusiveness}
\label{sec:inclusiveness}
\paragraph{\textbf{Posterior Probabilities}} Given two events A and B, the probability \(P(A|B)\) is said posterior probability because it allows to calculate the probability of A, knowing that B occurred. In our case the posterior probability means to compute the probability that \(Y = y\), knowing that \(A = a\) has occurred (and vice versa). In other words, the probability that the training set shows the property Y = y, knowing the property \(A = a\) has occurred (and vice versa). We start by estimating the probability that two events occur simultaneously. From the definition of conditional probability:
\begin{equation}\label{eq:condProb}
\begin{split}
    P(A = a\cap Y = y) = P(A=a)P(Y=y|A=a) \\
    P(Y = y\cap A = a) = P(Y=y)P(A=a|Y=y)
\end{split}
\end{equation}
Since from Compound Probability Theorem \cite{Ross:1996} \(P(A = a\cap Y = y)\) is equal to \(P(Y = y\cap A = a)\), i.e. the probability of both properties occurring is the same, either of the two formulas can be employed indistinctly.
\begin{table} [ht!]
\caption{Example of properties occurring simultaneously}
\label{tab:condProp}
\begin{tabular}{ll}
\toprule
\textbf{Formula} & \textbf{Probability}\\
\hline
\(P(Y = 0\cap A = white)\) & P = 0.42\\
\(P(Y = 0\cap A = black)\) & P = 0.07\\
\(P(Y = 0\cap A = Asian)\) & P = 0.09\\
\hline
\(P(Y = 1\cap A = white)\) & P = 0.18\\
\(P(Y = 1\cap A = black)\) & P = 0.28\\
\(P(Y = 1\cap A = Asian)\) & P = 0.06\\
\bottomrule
\end{tabular}%
\end{table}

\subsection{Estimating Training Likelihood}
\label{sec:likelihood}
From the definition of conditional probability, we derive the Bayes Theorem for the properties of the training set:
\begin{equation}\label{eq:bayes}
 \begin{split}
    P(A = a | Y = y) = \frac{P(A=a)P(Y=y|A=a)}{P(Y=y)} \\
    P(Y = y | A = a) = \frac{P(Y=y)P(A=a|Y=y)}{P(A=a)}
\end{split}   
\end{equation}
In the case of binary classification and in the case of protected attributes we are in the presence of a certain event partition. This means that the events are disjointed from each other \(Y_i \cap Y_j = \emptyset\) and \(A_i \cap A_j = \emptyset\) if \(i\neq j\) and that as a whole they are the only ones possible, i. e., if a certain property occurs, one and only one certainly appeared. In other words, it is not possible that the training set is composed of individuals who belong simultaneously to the black and white ethnic group, or who simultaneously show a positive and negative outcome. The union of the occurrence of the single properties is therefore the whole set of possible properties.
For the properties outcome and ethnicity the generalization formula are respectively:
\begin{equation}\label{eq:union}
\begin{split}
    \Omega: \cup_{i=1}^N Y_i = \Omega, \: hence \sum\limits_{i=1}^N P(Y_i) = P(\cup_{i=1}^N Y_i)\\
    \Omega: \cup_{i=1}^N A_i = \Omega, \: hence \sum\limits_{i=1}^N P(A_i) = P(\cup_{i=1}^N A_i)
\end{split}
\end{equation}
By applying Formulas \ref{eq:condProb} and \ref{eq:union} the Bayes Theorem can be generalized for each property of the training set:
\begin{equation}\label{eq:bayesGen}
\begin{split}
    P(Y = y | A) = \frac{P(Y=y)P(A|Y=y)}{P(A)} = \frac{P(Y=y)P(A|Y=y)}{\sum_{i=1}^N P(A|Y_i)P(Y_i)}\\
    P(A = a | Y) = \frac{P(A=a)P(Y|A=a)}{P(Y)} = \frac{P(A=a)P(Y|A=a)}{\sum_{i=1}^N P(Y|A_i)P(A_i)}
\end{split}    
\end{equation}
The first equation in Formula \ref{eq:bayesGen} derives the probability of the outcome property given the ethnic property, while the second equation derives the probability of the ethnic property given the outcome property. In other words, it derives the probability of composition of the training set based on the posterior probabilities of the outcome and ethnicity properties. Carried out a random sampling on the original data, the Formula answers the following questions:
\begin{itemize}
    \item [i] In the sampled training set what is the probability of belonging to an ethnic group with respect to the outcome variable?
    \item [ii] In the sampled training set what is the probability of obtaining a certain outcome with respect to the ethnic group?
\end{itemize}
Complementarily, the two equations can be interpreted as the probability of bias within the training set.
\begin{table} [H]
\caption{Example of posterior probabilities}
\label{tab:posterior}
\begin{tabular}{ll}
\toprule
\textbf{Formula} & \textbf{Probability}\\
\hline
\(P(Y = 0 | A = white)\) & P = 0.7\\
\(P(Y = 0 | A = black)\) & P = 0.2\\
\(P(Y = 0 | A = Asian)\) & P = 0.6\\
\hline
\(P(Y = 1 | A = white)\) & P = 0.3 \\
\(P(Y = 1 | A = black)\) & P = 0.8 \\
\(P(Y = 1 | A = Asian)\) & P = 0.4\\
\hline
\hline
\(P(A = white | Y = 1)\) & P = 0.34\\
\(P(A = white | Y = 0)\) & P = 0.87 \\
\hline
\(P(A = black | Y = 1)\) & P = 0.53\\
\(P(A = black | Y = 0)\) & P = 0.15\\
\hline
\(P(A = Asian | Y = 1)\) & P = 0.11\\
\(P(A = Asian | Y = 0)\) & P = 0.18\\
\bottomrule
\end{tabular}%
\end{table}

\section{Case studies datasets}
\label{sec:data}
\textbf{COMPAS} (Correctional Offender Management Profling for Alternative Sanctions)\footnote{Retrieved from: \url{https://www.propublica.org/datastore/dataset/compas-recidivism-risk-score-data-and-analysis}} is a popular tool used by U.S. court to estimate the defendants' probability of recidivism. This dataset displays the probability of reoffending based on two year of further studies. The dataset has been shown to underestimate the risk of recidivism for white defendants and overestimate it for black defendants \cite{Angwin:2016Compas}.\\
\textbf{Drug Consumption} \cite{Fehrman:2015data, Fehrman:2017paper} contains information on the consumption of 18 drugs based on personality traits and socio-economic attribute. For simplicity of analysis we assumed the consumption of Cannabis as target variable but the annotation of the dataset can be made on each target drug. \\
\textbf{Adult Dataset} \cite{Kohavi:1996} The data set contains adult income annual census from the US Census Bureau. It is commonly employed in forecasting tasks in order to predict the factors leading to income below or above \$50,000.

\begin{table} [H]
\caption{Summary of Datasets Prominent Properties}
\label{tab:dataProperties}
\begin{center}
\begin{tabular}{lrrr}
\toprule
\textbf{Property} & \multicolumn{1}{c}{\textbf{COMPAS}} & \multicolumn{1}{c}{\textbf{Drug}} & \multicolumn{1}{c}{\textbf{Adult}}\\
&  & \multicolumn{1}{c}{\textbf{Consumption}} & \multicolumn{1}{c}{\textbf{Dataset}}\\
\hline
Size & 6172x9 & 1885x31 & 48842x15 \\
\hline
Target  & 0 \(\to\) no & 0 \(\to\) non user& 0 \(\to > 50K\)\\
variable & 1 \(\to\) yes & 1 \(\to\) user & 1 \(\to \leq 50K\) \\
\hline
Levels of & Asian & Asian & AIE $^a$ \\
ethnicity & Black & Black & API $^b$\\
attribute & Caucasian & Black/Asian & Black \\
& Hispanic & Caucasian & Caucasian \\
& NA$^c$  & White/Asian & Other \\
& Other & White/Black & \\
&  & Other & \\
\bottomrule
\end{tabular}%
\end{center}
\footnotesize{$^a$ American-Indian/Eskimo, $^b$ Asian-Pac-Islander, $^c$ Native American}
\end{table}

\section{Results and Discussion}
\label{sec:results}
We performed the analyses that constitute our data annotation system for each of the datasets presented in Section \ref{sec:data}. Sub-sections \ref{sec:resDep}, \ref{sec:resDiv}, \ref{sec:resInc} and \ref{sec:resLik} report the analysis for each module - dependency, diverseness, inclusiveness, training likelihood, respectively - and contain an example graphic module. Figures \ref{Fig:compas} and \ref{Fig:compas1} shows an illustrative example of the graphical visualization for the complete notation. 

\subsection{Dependence}
\label{sec:resDep}
This module aims to analyze the connection relationships between the protected attribute Ethnicity and the target variable that are established and depend on the available data. For instance, for the COMPAS dataset the module highlights the dependency relationships between recidivism and different ethnic minorities. Summary results for dependence module are shown in Table \ref{tab:resDep}.
\begin{table} [H]
\caption{Summary of Dependence Prominent Properties}
\label{tab:resDep}
\begin{center}
\begin{tabular}{lccc}
\toprule
 & \multicolumn{1}{c}{\textbf{COMPAS}} & \multicolumn{1}{c}{\textbf{Drug}} & \multicolumn{1}{c}{\textbf{Adult}}\\
&  & \multicolumn{1}{c}{\textbf{Consumption}} & \multicolumn{1}{c}{\textbf{Dataset}}\\
\hline
Contingency & 0.1413 & 0.1558 & 0.0994\\
coefficient & & & \\
\hline
Effect size w & 0.1427 & 0.1578 & 0.0999\\
variable & & & \\
\hline
Magnitude of & SMALL & SMALL & VERY \\
Effect size w & & & SMALL \\
\bottomrule
\end{tabular}%
\end{center}
\end{table}
None of the three datasets displays worrying dependency values among the protected attribute Ethnicity and the target variable, showing the magnitude of the Effect Size w as small or very small. However, the results of the COMPAS dataset - which is proven to contain bias - indicate that this module alone is not sufficient to show a latent bias risk. The degree of bias depends on the sample size and the value of the contingency coefficient of the target variable and the protected attribute \cite{Zhou:2017}. Smaller samples lead to more bias and higher variance \cite{Zimmerman:2017} and therefore the results of the dependency must be analyzed in relation to the amount of data available. In order to facilitate the interpretation of the connection 
\begin{figure}[H]
  \includegraphics[width=0.28\textwidth]{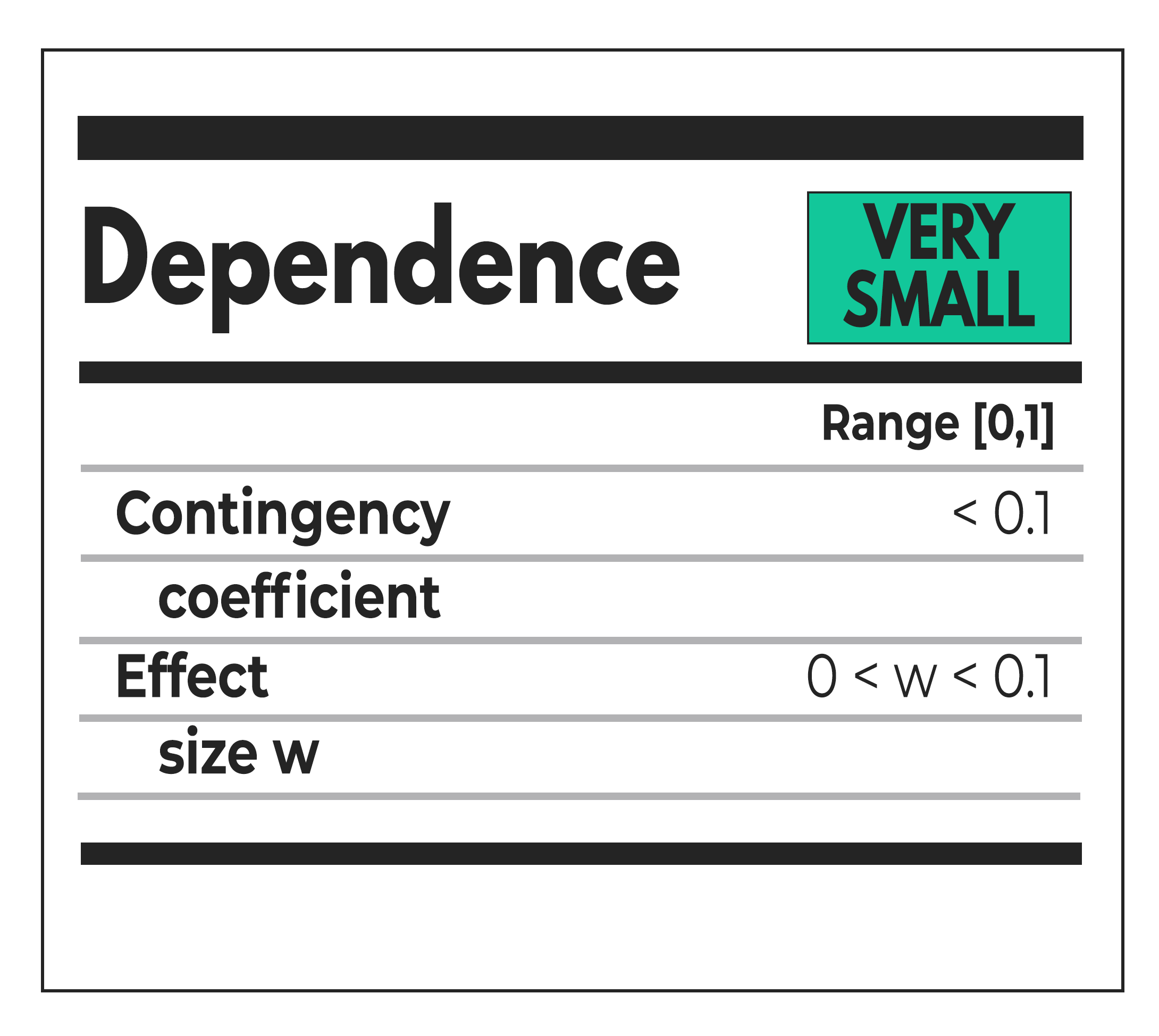}
  \centering
\end{figure}
\begin{figure}[H]
  \includegraphics[width=0.28\textwidth]{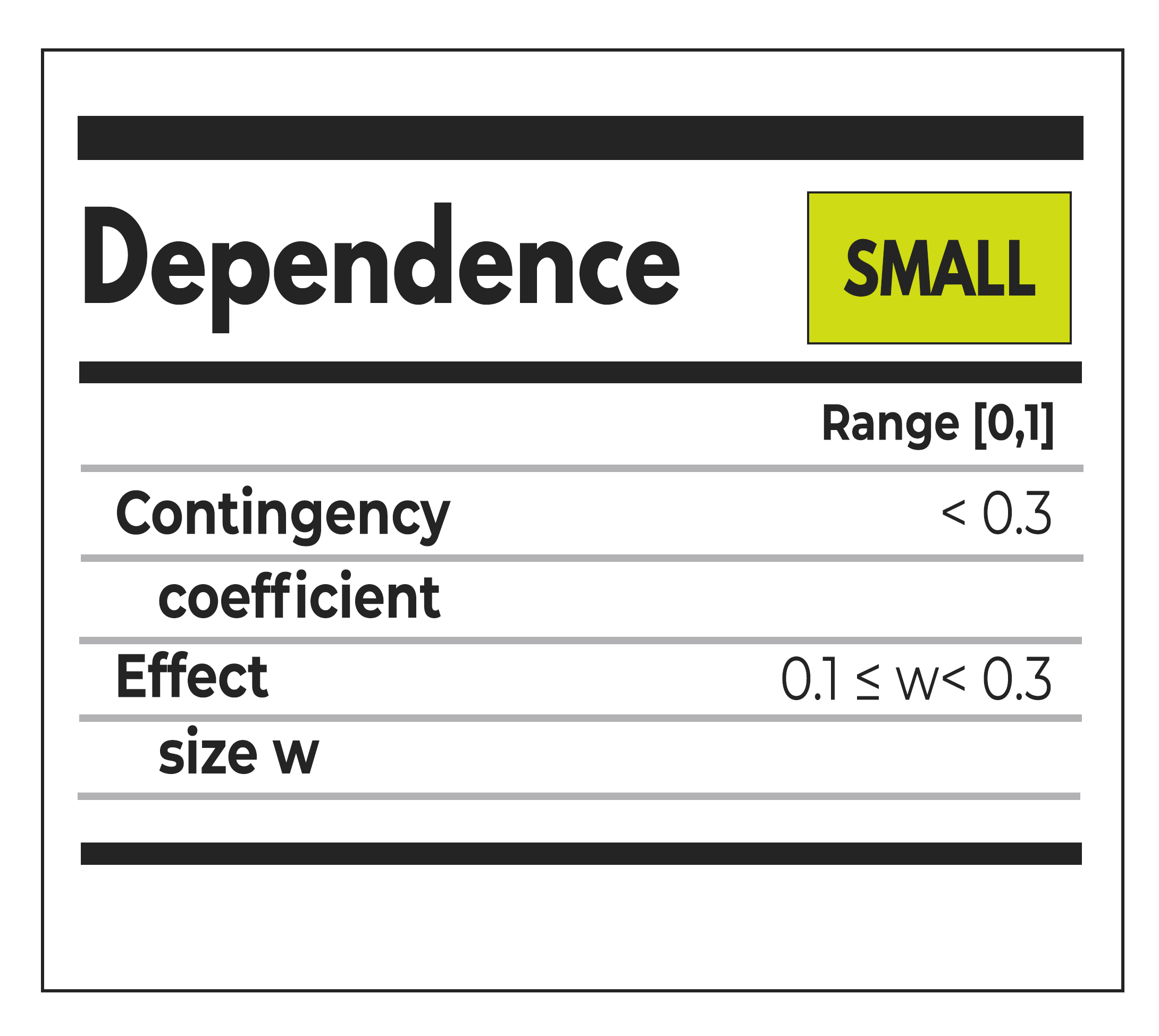}
  \centering
\end{figure}
\begin{figure}[H]
  \includegraphics[width=0.28\textwidth]{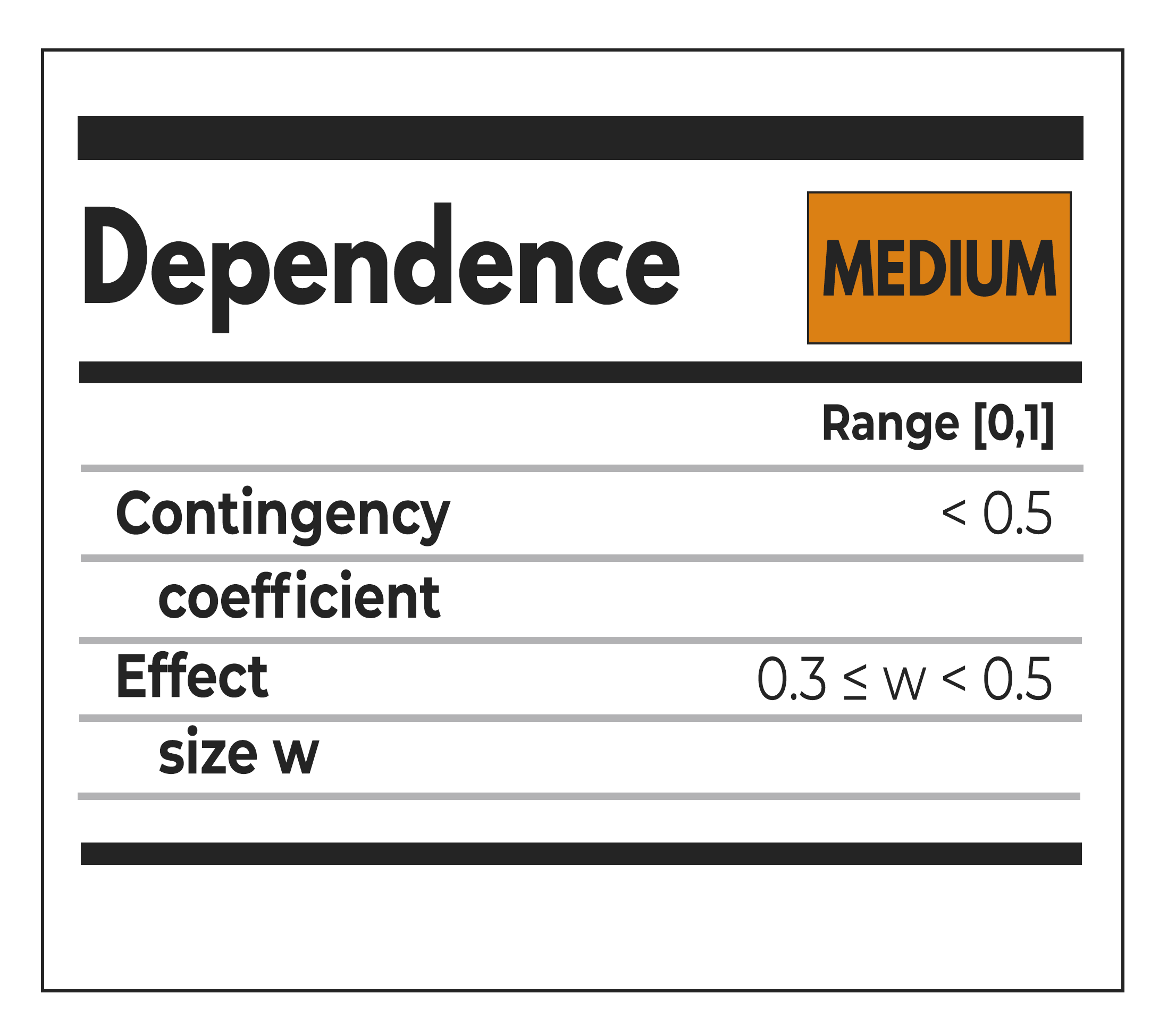}
  \centering
\end{figure}
\begin{figure}[H]
  \includegraphics[width=0.28\textwidth]{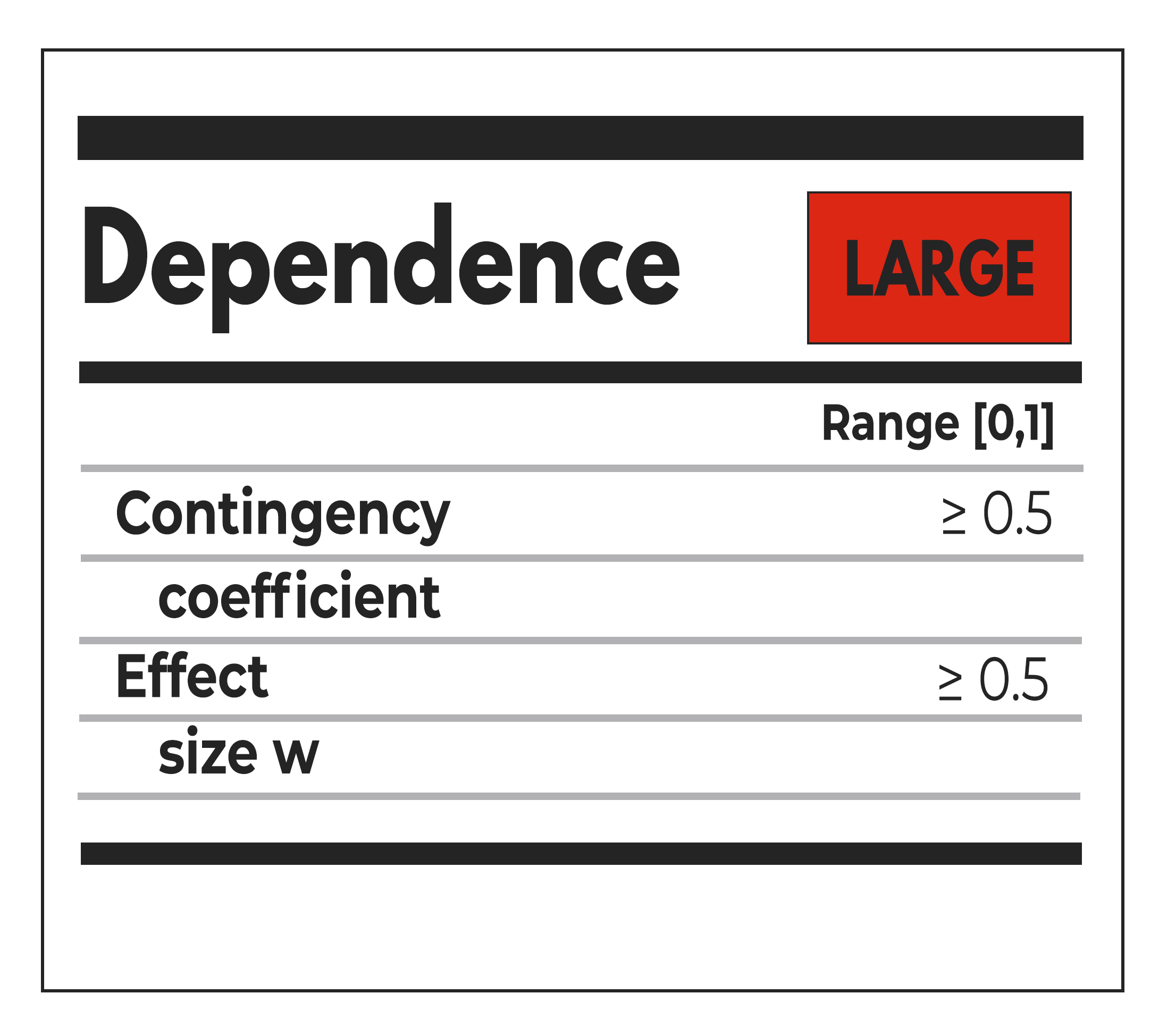}
  \centering
  \caption{Example of Dependence graphic visualization}
  \label{Fig:dependence}
\end{figure}
relations, we propose a graphic notation for dependence. 
Figure \ref{Fig:dependence} shows the graphical representations of the dependency modules based on different connection magnitude.

\subsection{Diverseness}
\label{sec:resDiv}
This module aims to analyze the diverseness of the data available by estimating prior probabilities. They determine the probability that training set will display an a priori environment based on the original data available, i.e. they show the probability of training set composition stratified by each of target variable and protected attribute levels.
For example, in our case study the module highlights the probability that training set will be equally composed by ethnic minorities and ethnic majorities.
Summary results for diverseness module are shown in Table \ref{tab:resDiv}.
\begin{table}
\caption{Summary of Diverseness Analysis Results}
\label{tab:resDiv}
\centering
\begin{tabular}{lrrr}
\toprule
& \multicolumn{1}{c}{\textbf{COMPAS}} & \multicolumn{1}{c}{\textbf{Drug}} & \multicolumn{1}{c}{\textbf{Adult}}\\
&  & \multicolumn{1}{c}{\textbf{Consumption}} & \multicolumn{1}{c}{\textbf{Dataset}}\\ 
  \hline
  0 & 0.545 & 0.329 & 0.239 \\ 
  1 & 0.455 & 0.671 & 0.761 \\
  \hline
  Caucasian & 0.341 & 0.912 & 0.855 \\ 
  Black & 0.514 & 0.018 & 0.096 \\ 
  Asian & 0.005 & 0.014 &  \\ 
  Hispanic & 0.082 &  &  \\ 
  Native American & 0.002 &  &  \\ 
  Other & 0.056 & 0.033 & 0.008 \\ 
  White/Black &  & 0.011 &  \\ 
  White/Asian &  & 0.011 &  \\ 
  Black/Asian &  & 0.002 &  \\ 
  Amer-Indian-Eskimo &  &  & 0.010 \\ 
  Asian-Pac-Islander &  &  & 0.031 \\
\bottomrule
\end{tabular}
\end{table}
In terms of target variable probabilities, the results show strong distortions for the Drug Consumption and Adult datasets with a high probability of positive examples - i.e. showing a negative outcome - while the probabilities of the COMPAS dataset are quite homogeneous. Regarding the probabilities of the protected attribute ethnicity, the distortions are even more pronounced than the target variable ones, revealing a very high probability of composition for the Caucasian ethnicity in the Drug Consumption and Adult datasets. In the case of the COMPAS dataset the probabilities are indeed distorted, although still not such as to predict at this point of the analysis more severe future distortions, which is why more in-depth analysis are required. Figure \ref{sec:diverseness} shows the graphical representation of the diverseness module that simplifies the display of prior probabilities. In the example is given the notation for a dataset where both the levels of the target variable and those of the protected attribute ethnicity are equiprobable.
\begin{figure}
  \includegraphics[width=0.25\textwidth]{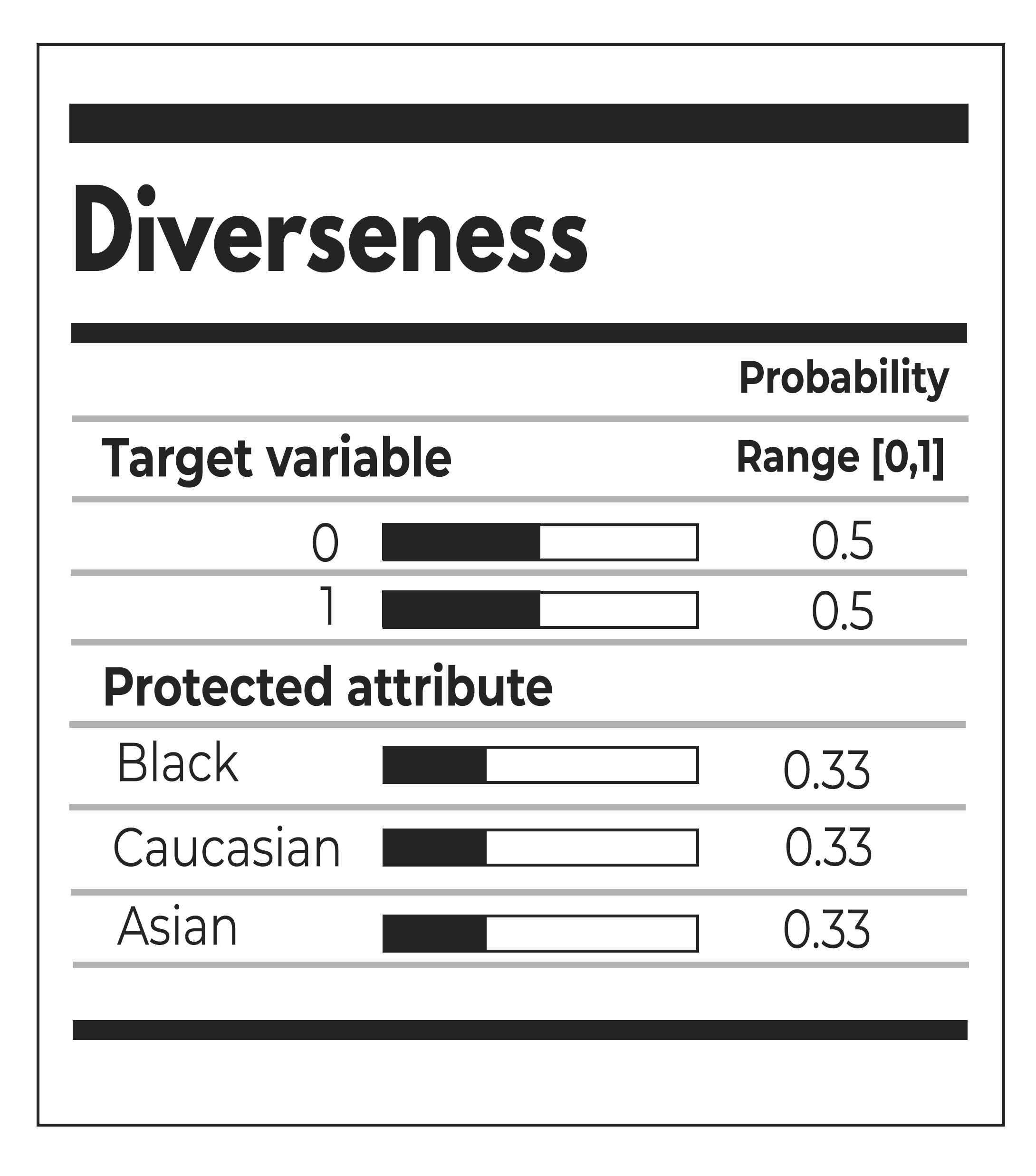}
  \centering
  \caption{Example of Diverseness graphic visualization}
  \label{Fig:div}
\end{figure}

\subsection{Inclusiveness}
\label{sec:resInc}
\begin{figure}
  \includegraphics[width=0.3\textwidth]{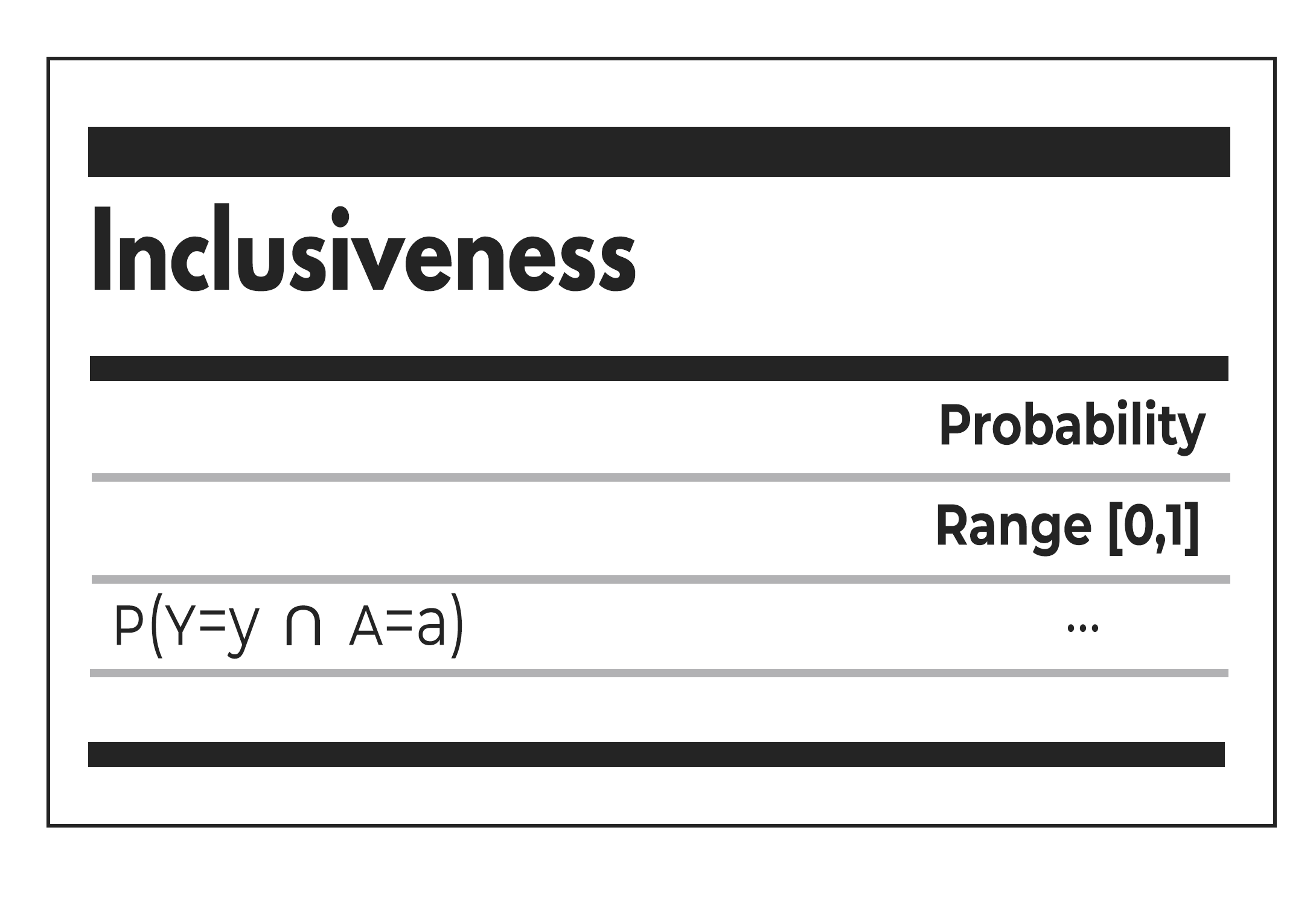}
  \centering
  \caption{Example of Inclusiveness graphic visualization}
  \label{Fig:Inclus}
\end{figure}

This module aims to analyze the inclusiveness of the data available by estimating the simultaneously probabilities. They determine the probability that training set will simultaneously display two by two the target variable and the protected attribute properties. For instance, in our case study the module highlights the probability that in training set the property Asian appears simultaneously with property success. Summary results for diverseness module are shown in Table \ref{tab:resInc}.
\begin{table}
\caption{Summary of Inclusiveness Analysis Results}
\label{tab:resInc}
\centering
\begin{center}
\begin{tabular}{lrrr}
\toprule
& \multicolumn{1}{c}{\textbf{COMPAS}} & \multicolumn{1}{c}{\textbf{Drug}} & \multicolumn{1}{c}{\textbf{Adult}}\\
&  & \multicolumn{1}{c}{\textbf{Consumption}} & \multicolumn{1}{c}{\textbf{Dataset}}\\ 
  \hline
  0\(\cap\)AIE$^a$ &  &  & 0.0006 \\ 
  0\(\cap\)Asian & 0.0023 & 0.0019 &  \\ 
  0\(\cap\)API$^b$ &  &  & 0.0041 \\ 
  0\(\cap\)Black & 0.1514 & 0.0023 & 0.0057 \\ 
  0\(\cap\)Black/Asian &  & 0.0000 &  \\ 
  0\(\cap\)Caucasian & 0.1281 & 0.0555 & 0.1061 \\ 
  0\(\cap\)Hispanic & 0.0320 &  &  \\ 
  0\(\cap\)NA$^c$ & 0.0006 &  &  \\ 
  0\(\cap\)Other & 0.0219 & 0.0013 & 0.0005 \\ 
  0\(\cap\)White/Asian &  & 0.0004 &  \\ 
  0\(\cap\)White/Black &  & 0.0006 &  \\ 
  \hline
  1\(\cap\)AIE &  &  & 0.0042 \\ 
  1\(\cap\)Asian & 0.0008 & 0.0007 &  \\ 
  1\(\cap\)API &  &  & 0.0111 \\ 
  1\(\cap\)Black & 0.1661 & 0.0010 & 0.0412 \\ 
  1\(\cap\)Black/Asian &  & 0.0003 &  \\ 
  1\(\cap\)Caucasian & 0.0822 & 0.1165 & 0.3115 \\ 
  1\(\cap\)Hispanic & 0.0189 &  &  \\ 
  1\(\cap\)NA & 0.0005 &  &  \\ 
  1\(\cap\)Other & 0.0124 & 0.0050 & 0.0036 \\ 
  1\(\cap\)White/Asian &  & 0.0016 &  \\ 
  1\(\cap\)White/Black &  & 0.0014 &  \\ 
\bottomrule
\end{tabular}
\end{center}
\footnotesize{$^a$ American-Indian/Eskimo, $^b$ Asian-Pac-Islander, $^c$ Native American}
\end{table}
The results of this module show that the probability that two properties will occur simultaneously is related to the sample size. Evidence of this can be found in the results of the Drug Consumption and Adult datasets, where the highest probabilities of simultaneous events involve the Caucasian property. The COMPAS dataset shows quite homogeneous probabilities especially with regard to the Black property, while for the Caucasian property the highest probabilities are related to the simultaneous occurrence with the Non-recidivist property. 
Since the simultaneous probabilities depend on the number of examples within the available data and the sample size, this result alone is not sufficient to establish a priori the certain presence of serious data distortions, although some evidence can already be seen. Figure \ref{Fig:Inclus} shows the graphical representation of the inclusiveness module that simplifies the display of simultaneously probabilities. In the example is given the notation for a dataset where all the properties of the target variable and those of the protected attribute ethnicity are equiprobable.

\subsection{Training Likelihood}
\label{sec:resLik}
This module aims to analyze the training likelihood of the data available by estimating the posterior probabilities. They determine the probability that in the training set the occurrence of the properties of the protected attribute is given by the properties of the target variable - and vice versa - . For example, in the COMPAS dataset they determine the probability that the occurrence of reoffending is given by the properties of the protected attribute ethnicity. Summary results for training likelihood module are shown in Table \ref{tab:resLik}.

\begin{figure}
  \includegraphics[width=0.3\textwidth]{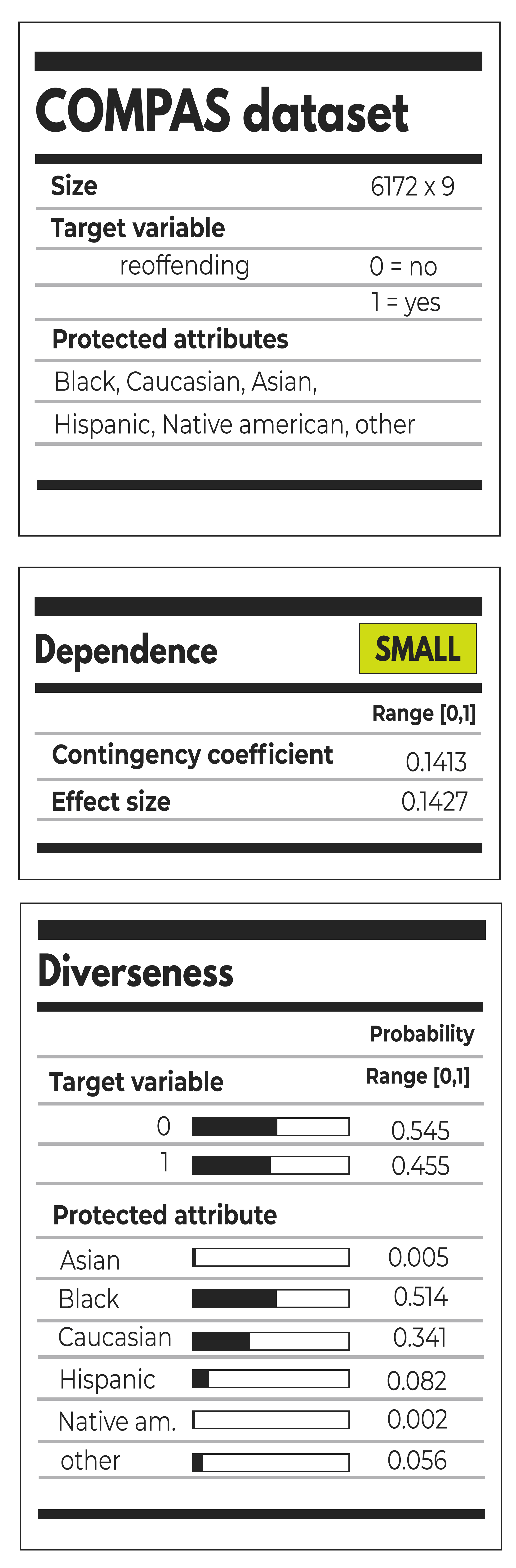}
  \centering
  \caption{Data annotation visualization for COMPAS dataset}
  \label{Fig:compas}
\end{figure}

\begin{figure}
  \includegraphics[width=0.3\textwidth]{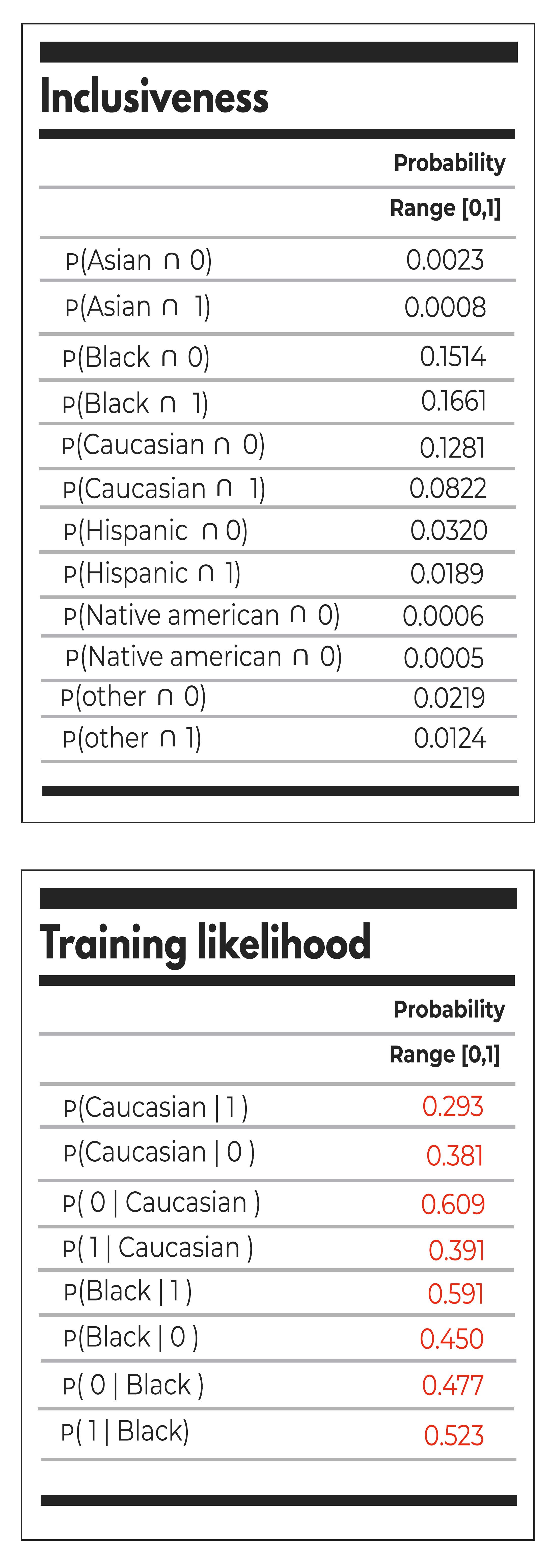}
  \centering
  \caption{Data annotation visualization for COMPAS dataset}
  \label{Fig:compas1}
\end{figure}

\begin{table}
\caption{Summary of Training Likelihood Analysis Results}
\label{tab:resLik}
\centering
\begin{center}
\begin{tabular}{lrrr}
\toprule
& \multicolumn{1}{c}{\textbf{COMPAS}} & \multicolumn{1}{c}{\textbf{Drug}} & \multicolumn{1}{c}{\textbf{Adult}}\\
&  & \multicolumn{1}{c}{\textbf{Consumption}} & \multicolumn{1}{c}{\textbf{Dataset}}\\
  \hline
  0\(|\)AIE$^a$ &  &  & 0.117 \\ 
  0\(|\)Asian & 0.742 & 0.731 &  \\ 
  0\(|\)API$^b$ &  &  & 0.269 \\ 
  0\(|\)Black & 0.477 & 0.697 & 0.121 \\ 
  0\(|\)Black/Asian &  & 0.000 &  \\ 
  0\(|\)Caucasian & 0.609 & 0.323 & 0.254 \\ 
  0\(|\)Hispanic & 0.629 &  &  \\ 
  0\(|\)NA$^c$ & 0.545 &  &  \\ 
  0\(|\)Other & 0.638 & 0.206 & 0.123 \\ 
  0\(|\)White/Asian &  & 0.200 &  \\ 
  0\(|\)White/Black &  & 0.300 &  \\ 
  1\(|\)AIE &  &  & 0.883 \\ 
  1\(|\)Asian & 0.258 & 0.269 &  \\ 
  1\(|\)API &  &  & 0.731 \\ 
  1\(|\)Black & 0.523 & 0.303 & 0.879 \\ 
  1\(|\)Black/Asian &  & 1.000 &  \\ 
  1\(|\)Caucasian & 0.391 & 0.677 & 0.746 \\ 
  1\(|\)Hispanic & 0.371 &  &  \\ 
  1\(|\)NA & 0.455 &  &  \\ 
  1\(|\)Other & 0.362 & 0.794 & 0.877 \\ 
  1\(|\)White/Asian &  & 0.800 &  \\ 
  1\(|\)White/Black &  & 0.700 &  \\ 
  \hline
  AIE\(|\)0 &  &  & 0.005 \\ 
  AIE\(|\)1 &  &  & 0.011 \\ 
  Asian\(|\)0 & 0.007 & 0.031 &  \\ 
  Asian\(|\)1 & 0.003 & 0.006 &  \\ 
  API\(|\)0 &  &  & 0.035 \\ 
  API\(|\)1 &  &  & 0.030 \\ 
  Black\(|\)0 & 0.450 & 0.037 & 0.048 \\ 
  Black\(|\)1 & 0.591 & 0.008 & 0.111 \\ 
  Black/Asian\(|\)0 &  & 0.000 &  \\ 
  Black/Asian\(|\)1 &  & 0.002 &  \\ 
  Caucasian\(|\)0 & 0.381 & 0.895 & 0.908 \\ 
  Caucasian\(|\)1 & 0.293 & 0.921 & 0.839 \\ 
  Hispanic\(|\)0 & 0.095 &  &  \\ 
  Hispanic\(|\)1 & 0.067 &  &  \\ 
  NA\(|\)0 & 0.002 &  &  \\ 
  NA\(|\)1 & 0.002 &  &  \\ 
  Other\(|\)0 & 0.065 & 0.021 & 0.004 \\ 
  Other\(|\)1 & 0.044 & 0.040 & 0.010 \\ 
  White/Asian\(|\)0 &  & 0.006 &  \\ 
  White/Asian\(|\)1 &  & 0.013 &  \\ 
  White/Black\(|\)0 &  & 0.010 &  \\ 
  White/Black\(|\)1 &  & 0.011 &  \\ 
\bottomrule
\end{tabular}
\end{center}
\footnotesize{$^a$ American-Indian/Eskimo, $^b$ Asian-Pac-Islander, $^c$ Native American}
\end{table}
The results of this module show that the posterior probabilities of target variable and protected attribute ethnicity are quite skewed in all dataset. In the case of the Adult dataset given as occurred event 1 or event 0, the probability of occurrence of the Caucasian ethnic group is respectively 0.908 and 0.839, - i.e. very high for both events - while the probabilities of all other ethnic groups conditioned to the target variable are all significantly lower; this means that the original data contain many examples of individuals belonging to the Caucasian ethnic group. In the case of Drug Consumption, a similar reasoning can be carried out for the ethnicity probabilities conditioned to the target variable; moreover, notice that given the property Black/Asian, the probability of occurrence of event 1, i. e. that the individual is a consumer, is 1 - while the probability of 0 is 0 - which means that in the available data there are no examples of individuals belonging to the ethnic group Black/Asian showing a positive outcome - i. e. negative examples -. 
Figures \ref{Fig:compas} and \ref{Fig:compas1} shows the graphical visualization of our data annotation system for the COMPAS dataset.
The analysis of the COMPAS dataset shows that if an individual is randomly sampled from the original data for the training set, the probability that this individual is black knowing that the re-offending property has occurred - i.e. knowing the outcome of the re-offending event - is 0.591, while the probability that the individual is white knowing that the re-offending property has occurred is 0.293. Instead, given as occurred the property Black the probability that the individual has not reoffended is 0.477, while the probability that the individual has reoffended is 0.523; given the property Caucasian, the probability that the individual has not reoffended is 0.609, while the probability that the individual has reoffended is 0.391, that is significantly lower.
This means that in this dataset the reoffending is related to ethnicity, and that success or failure are determined by the membership to a specific ethnic group. The differences in probability between the properties highlight the risk of future bias, and in the case of the COMPAS dataset they anticipate the underestimation of recidivism for the Caucasian ethnic group and the overestimation of recidivism for the Black ethnic group proven in recent studies \cite{Angwin:2016Compas}.

\subsection{Final Remarks}
\label{sec:FR}
\begin{itemize}
    \item [\textbf{RQ1}:]in traditional sampling practices, instead of observing all the units of a population, only a subset of a population is detected, which must show certain probabilistic characteristics. In machine learning models the training set is sampled not from the real population but from the available data. While in classical sampling the empirical knowledge alone is effectively of a sample nature, in machine learning systems the available data are often of sample nature too, precisely due to the fact that it is not possible to make assumptions on the real population. Considering a random sampling from the available data, we have shown that the probability of composition of the training set can be predicted, highlighting that the structure of the data directly affects the probability of properties distribution;
    \item[\textbf{RQ2}:] we analyzed three datasets frequently accessed by machine learning community. Of these, all three showed more or less pronounced distortions for the protected attribute Ethnicity. Although the COMPAS dataset is the sole one that has been shown to discriminate against black people, the Drug Compsuntion and Adult datasets reveal possible future bias in the detriment of ethnic minorities.
\end{itemize}

\section{Related Work}
\label{sec:relatedWork}
Although there are a number of papers that for ethical purposes deal with data annotation they are all very recent, indicating that this field of study is still partially explored and has only recently received considerable attention. Our contribution differs from the others because it induces a probabilistic reasoning on the causes of model discrimination based on sampling problems; our intention is to deepen the knowledge of data validation analysis, focusing on the meaning of probabilities. From a graphical point of view, our work has been inspired by the Data Nutrition Labels \cite{Holland:2018}, a data labeling system mainly based on descriptive data statistics. A similar approach is addressed in \cite{Beretta:19}, where an operational framework is proposed to identify the bias risks of automatic decision systems. In \cite{Gebru:2018} the authors propose a data labeling system based on discursive data sheets. In \cite{Chang:2017} the authors propose a collaborative crowdsourcing system to improve the quality of the labels.

Since ethically data annotation represent a quite new field of study, there are several works that provide different types of labels. We believe that at present the focus should not be on achieving a unified data annotation system in the short term, but rather on the fact that the fair machine learning community is working together to focus attention on the data collection problem. Especially because awareness of data issues is often not rooted outside of this community. It is important that this field and this work inspire greater awareness of the possible causes of discrimination due to the fundamental ingredient that all users and designers of machine learning systems (from the most to the least experienced) use, data.

\section{Conclusions}
\label{sec:conclusions}
The purpose of the current study was to detect the potential race discriminatory risk for future machine learning system by providing a data annotation system based on Bayesian Inference. Our notation serves as a diagnostic framework to immediately visualize data appropriateness and potential bias occurring when sampling the training set from an available dataset. 
The investigation of the probabilities of the training set sampling has shown that it is possible to establish a risk of future bias by observing prior and posterior probabilities of the ethnicity and target variable properties. The empirical findings in this study provide a new perspective on data annotation practices by showing that Bayesian inferences may reveal the risk of bias in three different widespread dataset. Furthermore, this study has raised important questions about the awareness of most widely data sampling practices in machine learning community. The findings of this investigation complement those of earlier studies. Our data annotation system is limited to the binary case and to the analysis of categorical variables for classification tasks. This would be a fruitful area for further work. Our intent is to expand the work in the following directions: i) extend the notation to multiple protected attributes - the probabilities of the training set will then be given by the vectors of the protected attribute combinations - ; ii) extend the notation to the non-binary case - for prediction tasks involving regression analysis for example - ; iii) extend the probabilistic notation to non-labeled data.

\bibliographystyle{ACM-Reference-Format}

\end{document}